# Inaccuracy Minimization by Partitioning Fuzzy Data Sets – Validation of an Analytical Methodology

Arutchelvan.G
 Department of Computer Science and Applications
Adhiparasakthi College of Arts and Science
G.B.Nagar, Kalavai , India
garutchelvan@yahoo.com

Dr.Sivatsa S.K
Professor (Retired),
Dept. of Electronics Engineering,
Madras Institute of Technology,
Anna University,Chennai, India
profsks@rediffmail.com

Dr. Jaganathan. R
Deputy Registrar(Edn)
Vinayaka Mission
University, Chennai,
India,
Jaganr1@yahoo.com

*Abstract*— **In the last two decades, a number of methods have been proposed for forecasting based on fuzzy time series. Most of the fuzzy time series methods are presented for forecasting of car road accidents. However , the forecasting accuracy rates of the existing methods are not good enough. In this paper, we compared our proposed new method of fuzzy time series forecasting with existing methods. Our method is based on means based partitioning of the historical data of car road accidents. The proposed method belongs to the kth order and time-variant methods. The proposed method can get the best forecasting accuracy rate for forecasting the car road accidents than the existing methods.**

## I. INTRODUCTION

Forecasting activities play an important role in our daily life. During last two decades, various approaches have been developed for time series forecasting. Among them, ARMA models and Box-Jenkins model building approaches are highly famous. However, the classical time series methods can not deal with forecasting problems in which the values of time series are linguistic terms represented by fuzzy sets [11], [23]. Therefore, Song and Chissom [18] presented the theory of fuzzy time series to overcome this drawback of the classical time series methods. Based on the theory of fuzzy time series, Song et al. presented some forecasting methods [16], [18], [19], [20] to forecast the enrollments of the University of Alabama. In [1] Chen and Hsu and in [2], Chen presented a method to forecast the enrollments of the University of Alabama based on fuzzy times series. It has the advantage of reducing the calculation, time and simplifying the calculation process. In [8], Hwang, Chen and Lee used the differences of the enrollments to present a method to forecast the enrollments of the University of Alabama based on fuzzy time series. In [5] and [6], Huang used simplified calculations with the addition of heuristic rules to forecast the enrollments using [2].

In [4], Chen presented a forecasting methods based on high-order fuzzy time series for forecasting the enrollments of the University of Alabama. In [3], Chen and Hwang presented method based on fuzzy time series to forecast the daily temperature. In [15], Melike and Konstsntin presented a new first order time series model for forecasting enrollments of the University of Alabama. In [14], Li and Kozma presented a dynamic neural network method for time series prediction using the KIII model. In [21], Su and Li presented a method for fusing global and local information in predicting time series based on neural networks. In [22], Sullivan and Woodall reviewed the first-order time-variant fuzzy time series model and the first-order time-invariant fuzzy time series model presented by Song and Chissom[18], where their models are compared with each other and with a time-variant Markov model using linguistic labels with probability distributions. In [13], Lee, Wang and Chen presented two factor high order fuzzy time series for forecasting daily temperature in Taipei and TAIFEX. In [9], Jilani and Burney and in [10], Jilani, Burney and Ardil presented new fuzzy metrics for high order multivariate fuzzy time series forecasting for car road accident casualties in Belgium.

   In this paper, we present a comparison of our proposed method and existing fuzzy time series forecasting methods to forecast the car road accidents in Belgium. Our proposed method belongs to the class of k-step first-order univariate time-variant method. The proposed method gives best forecasting accuracy rate for forecasting the car road accidents when compared with existing methods. The rest of this paper is organized as follows. In section2, we briefly review some basic concepts of fuzzy time series. In Section 3, we present our method of fuzzy forecasting based on means of partitioning the car road accidents data. In Section 4, we compared the forecasting results of the proposed method with the existing methods. In section 5, we conclude the results.

## II BASIC CONCEPTS OF FUZZY TIME SERIES:

There are number of definitions for fuzzy time series.

**Definition 1:** Imprecise data at equally spaced discrete time points are modeled as fuzzy variables. The set of this discrete fuzzy data forms a fuzzy time series.

**Definition 2:** Chronological sequences of imprecise data are considered as time series with fuzzy data. A time series with fuzzy data is refereed to as fuzzy time series.

**Definition 3:** Let $Y(t)$, $(t=...0,1,2,...)$ be the universe of discourse and $Y(t) \delta R$. Assume that $f_i(t)$, $i=1,2,...$ is defined in the universe of discourse $Y(t)$ and $F(t)$ is a collection of $f(t_i)$, ($i=...0,1,2,...$), then $F(t)$ is called a fuzzy time series of $Y(t)$, $i=1,2,..$ Using fuzzy relation, we define $F(t)=F(t-1) \circ R(t,t-1)$ where $R(t,t-1)$ is a fuzzy relation and "o" is the max min composition operator, then $F(t)$ is caused by $F(t-1)$ where $F(t)$ and $F(t-1)$ are fuzzy sets. Let $F(t)$ be a fuzzy time series and let $R(t, t-1)$ be a first order model of $F(t)$. If $R(t, t-1)=R(t-1,t-2)$ for any time t then $F(t)$ is called a time-invariant fuzzy time series. If $R(t,t-1)$ is dependent on time t, that is, $R(t, t-1)$ may be different from $R(t-1,t-2)$ for any t, then $F(t)$ is called a time-variant fuzzy time series.

## III. PROPOSED METHOD USING FUZZY TIME SERIES:

In this section, we present our method to forecast the data of car road accidents based on means. The historical data of car road accidents are given in Table I. First, based on [1], we defined the partition the universe of discourse into equal length intervals. Then based on means of original data and means of frequency of interval data, we redefine the intervals. After this, define some membership function for each interval of the historical data to obtain fuzzy data to form a fuzzy time series. Then, it establishes fuzzy logical relationships (FLRs) based on the fuzzified data in Table IV. Finally, it uses our proposed method.

**Step 1:** Define the universe of discourse U and partition it into four equal intervals $u_1, u_2, u_3, \ldots u_n$. For example, assume that the universe of discourse $U=[900, 17000]$ is partitioned into four even and equal length intervals.

**Step 2:** Get a mean of the original data. Get the means of frequency of each interval. Compare the means of original and frequency of each interval and then split the four interval into number of sub-interval respectively.

**Step 3:** Define each fuzzy set $A_i$ based on the re-divided intervals and fuzzify the historical data shown in Table I, where fuzzy set $A_i$ denotes a linguistic value of the accident data represented by a fuzzy set. We have used triangular membership function to define the fuzzy sets $A_i$ [10]. The reason for fuzzifying the accident data into fuzzified data is to translate crisp values into fuzzy sets to get a fuzzy time series.

**Step 4:** Establish fuzzy logical relationships based on the fuzzified data where the fuzzy logical relationship "$A_p$, $A_q$, $A_r$ -> $A_s$" denotes that "if the fuzzified data of year p, q and r are $A_p$, $A_q$ and $A_r$ respectively, then the fuzzified data of the year ( r ) is $A_r$".

$$t_j = \begin{cases} \dfrac{1+0.5}{(1/a_1)+(0.5/a_2)} & \text{if } j=1, \\[2ex] \dfrac{0.5+1+0.5}{(0.5/a_{j-1})+(1/a_j)+(0.5/a_{j+1})} & \text{if } 2 \leq j \leq n-2 \\[2ex] \dfrac{0.5+1}{(0.5/a_{n-1})+(1/a_n)} & \text{if } j=n \end{cases}$$

where $a_{j-1}, a_j, a_{j+1}$ are the mid points of the fuzzy intervals $A_{j-1}, A_j, A_{j+1}$ respectively. Based on the fuzzify historical enrollments obtained in step 3, we can get the fuzzy logical relationship group (FLGR) as shown in Table IV.

Divide each interval derived in step 2 into subintervals of equal length with respect to the corresponding means of the interval. We have assumed twenty nine partitions of the universe of discourse of the main factor fuzzy time series. Assuming that $0 \neq A_i$, for every $A_i$, $i=1,2,\ldots 29$.

TABLE – I

**YEARLY CAR ACCIDENTS FROM 1974 TO 2004 IN BELGIUM**

| Year | Killed | Year | Killed |
|------|--------|------|--------|
| 2004 | 953    | 1984 | 1,369  |
| 2003 | 1,035  | 1983 | 1,479  |
| 2002 | 1,145  | 1982 | 1,464  |
| 2001 | 1,288  | 1981 | 1,564  |
| 2000 | 1,253  | 1980 | 1,616  |
| 1999 | 1,173  | 1979 | 1,572  |
| 1998 | 1,224  | 1978 | 1,644  |
| 1997 | 1,150  | 1977 | 1,597  |
| 1996 | 1,122  | 1976 | 1,536  |
| 1995 | 1,228  | 1975 | 1,460  |
| 1994 | 1,415  | 1974 | 1,574  |
| 1993 | 1,346  |      |        |
| 1992 | 1,380  |      |        |
| 1991 | 1,471  |      |        |
| 1990 | 1,574  |      |        |
| 1989 | 1,488  |      |        |
| 1988 | 1,432  |      |        |
| 1987 | 1,390  |      |        |
| 1986 | 1,456  |      |        |
| 1985 | 1,308  |      |        |

TABLE II
**Means of Original Data and frequency of intervals data**

| Interval | Re-divide the interval |
|---|---|
| [900, 1100] | 1 |
| [1100, 1300] | 6 |
| [1300, 1500] | 13 |
| [1500, 1700] | 9 |

TABLE III

**FUZZY INTERVALS USING MEAN BASED PARTITIONING**

| Linguistics | Intervals |
|---|---|
| $u_1$ | [900, 1100] |
| $u_2$ | [1100, 1133.33] |
| $u_3$ | [1133.33, 1166.66] |
| $u_4$ | [1166.66, 1199.99] |
| $u_5$ | [1199.99, 1233.32] |
| $u_6$ | [1233.32, 1266.65] |
| $u_7$ | [1266.65, 1300.00] |
| $u_8$ | [1300.00, 1315.38] |
| $u_9$ | [1315.38, 1330.76] |
| $u_{10}$ | [1330.76, 1346.14] |
| $u_{11}$ | [1346.14, 1361.52] |
| $u_{12}$ | [1361.52, 1376.90] |
| $u_{13}$ | [1376.90, 1392.28] |
| $u_{14}$ | [1392.28, 1407.66] |
| $u_{15}$ | [1407.66, 1423.04] |
| $u_{16}$ | [1423.04, 1438.42] |
| $u_{17}$ | [1438.42, 1453.80] |
| $u_{18}$ | [1453.80, 1469.18] |
| $u_{19}$ | [1469.18, 1484.56] |
| $u_{20}$ | [1484.56, 1500.00] |
| $u_{21}$ | [1500.00, 1522.22] |
| $u_{22}$ | [1522.22, 1544.44] |
| $u_{23}$ | [1544.44, 1566.66] |
| $u_{24}$ | [1566.66, 1588.88] |
| $u_{25}$ | [1588.88, 1611.10] |
| $u_{26}$ | [1611.10, 1633.32] |
| $u_{27}$ | [1633.32, 1655.54] |
| $u_{28}$ | [1655.54, 1677.76] |
| $u_{29}$ | [1677.76, 1700.00] |

### IV A COMPARISON OF DIFFERENT FORECASTING METHODS

In the following Table V summarizes the forecasting results of the proposed method from 1974 to 2004, where the universe of discourse is divided into twenty nine intervals based on means based partitioning. In the following, we use the average forecasting error rate (AFER) and mean square error (MSE) to compare the forecasting results of different forecasting methods, where $A_i$ denotes the actual value and $F_i$ denotes the forecasting value of year (i), respectively.

$$AFER = \frac{|A_i - F_i|/A_i}{n} * 100\%$$

$$MSE = \frac{\sum_{i=1}^{n}(A_i - F_i)^2}{n}$$

In Table VI, we compare the forecasting results of the proposed method with that of the existing methods. From Table III, we can see that when the number of intervals in the universe of discourse is twenty nine intervals are sub partitioned base on means, the proposed method shows smallest values of the MSE and AFER of the forecasting results as compared to other methods of fuzzy time series forecasting. That is, the proposed method can get a higher forecasting accuracy rate for forecasting car road accidents that the existing methods.

TABLE IV

**THIRD ORDER FUZZY LOGICAL RELATIONSHIP GROUPS**

Group 1: $X_1, X_3, X_7 \rightarrow X_6$

Group 2: $X_3, X_7, X_6 \rightarrow X_4$

Group 3: $X_7, X_6, X_4 \rightarrow X_5$

Group 4: $X_6, X_4, X_5 \rightarrow X_3$

Group 5: $X_4, X_5, X_3 \rightarrow X_2$

Group 6: $X_5, X_3, X_2 \rightarrow X_5$

Group 7: $X_3, X_2, X_5 \rightarrow X_{15}$

Group 8: $X_2, X_5, X_{15} \rightarrow X_{10}$

Group 9: $X_5, X_{15}, X_{10} \rightarrow X_{13}$

Group 10: $X_{15}, X_{10}, X_{13} \rightarrow X_{19}$

Group 11: $X_{10}, X_{13}, X_{19} \rightarrow X_{24}$

Group 12: $X_{13}, X_{19}, X_{24} \rightarrow X_{20}$

Group 13: $X_{19}, X_{24}, X_{20} \rightarrow X_{16}$

Group 14: $X_{24}, X_{20}, X_{16} \rightarrow X_{13}$

Group 15: $X_{20}, X_{16}, X_{13} \rightarrow X_{18}$

Group 16: $X_{16}, X_{13}, X_{18} \rightarrow X_8$

Group 17: $X_{13}, X_{18}, X_8 \rightarrow X_{12}$

Group 18: $X_{18}, X_{13}, X_{12} \rightarrow X_{19}$

Group 19: $X_{13}, x_{12}, x_{19}, \rightarrow X_{18}$

Group 20: $X_{12}, X_{19}, X_{18} \rightarrow X_{23}$

Group 21: $X_{19}, X_{18}, X_{23} \rightarrow X_{26}$

Group 22: $X_{18}, X_{23}, X_{26} \rightarrow X_{24}$

Group 23: $X_{23}, X_{26}, X_{24} \rightarrow X_{27}$

Group 24: $X_{26}, X_{24}, X_{27} \rightarrow X_{22}$

Group 25: $x_{24}, X_{27}, X_{22} \rightarrow X_{18}$

Group 26: $X_{27}, X_{22}, X_{18} \rightarrow X_{24}$

**TABLE -V**

| Years | Actual Values $A_i$ | Mid Values of Interval | Calculated Values $F_i$ | $(A_i-F_i)^2$ | $\|A_i - F_i\| / A_i$ |
|---|---|---|---|---|---|
| 2004 | 953 | 1,000.0000 | 1,036.0825 | 6,902.7051 | 0.0872 |
| 2003 | 1035 | 1,000.0000 | 1,036.0825 | 1.1718 | 0.0010 |
| 2002 | 1145 | 1,150.0000 | 1,149.5167 | 20.4008 | 0.0039 |
| 2001 | 1288 | 1,283.3335 | 1,280.7594 | 52.4263 | 0.0056 |
| 2000 | 1253 | 1,250.0002 | 1,249.5555 | 11.8643 | 0.0027 |
| 1999 | 1173 | 1,183.3335 | 1,182.8638 | 97.2940 | 0.0084 |
| 1998 | 1224 | 1,216.6667 | 1,216.2100 | 60.6847 | 0.0064 |
| 1997 | 1150 | 1,150.0000 | 1,149.5167 | 0.2336 | 0.0004 |
| 1996 | 1122 | 1,116.6667 | 1,092.7141 | 857.6633 | 0.0261 |
| 1995 | 1228 | 1,216.6667 | 1,216.2100 | 139.0050 | 0.0096 |
| 1994 | 1415 | 1,415.3848 | 1,415.3013 | 0.0908 | 0.0002 |
| 1993 | 1346 | 1,338.4617 | 1,338.3732 | 58.1686 | 0.0057 |
| 1992 | 1380 | 1,384.6155 | 1,384.5300 | 20.5212 | 0.0033 |
| 1991 | 1471 | 1,476.9233 | 1,476.8433 | 34.1437 | 0.0040 |
| 1990 | 1574 | 1,577.7776 | 1,577.6211 | 13.1123 | 0.0023 |
| 1989 | 1488 | 1,492.3079 | 1,493.0643 | 25.6474 | 0.0034 |
| 1988 | 1432 | 1,430.7695 | 1,430.6868 | 1.7246 | 0.0009 |
| 1987 | 1390 | 1,384.6155 | 1,384.5300 | 29.9206 | 0.0039 |
| 1986 | 1456 | 1,461.5388 | 1,461.4578 | 29.7872 | 0.0037 |
| 1985 | 1308 | 1,307.6924 | 1,305.2928 | 7.3287 | 0.0021 |
| 1984 | 1369 | 1,369.2310 | 1,369.1444 | 0.0209 | 0.0001 |
| 1983 | 1479 | 1,476.9233 | 1,476.8433 | 4.6515 | 0.0015 |
| 1982 | 1464 | 1,461.5388 | 1,461.4578 | 6.4630 | 0.0017 |
| 1981 | 1564 | 1,555.5554 | 1,555.3967 | 74.0163 | 0.0055 |
| 1980 | 1616 | 1,622.2219 | 1,622.0697 | 36.8413 | 0.0038 |
| 1979 | 1572 | 1,577.7776 | 1,577.6211 | 31.5967 | 0.0036 |
| 1978 | 1644 | 1,644.4441 | 1,644.2939 | 0.0864 | 0.0002 |
| 1977 | 1597 | 1,599.9998 | 1,599.8455 | 8.0966 | 0.0018 |
| 1976 | 1536 | 1,533.3333 | 1,533.1722 | 7.9962 | 0.0018 |
| 1975 | 1460 | 1,461.5388 | 1,461.4578 | 2.1251 | 0.0010 |
| 1974 | 1574 | 1,577.7776 | 1,577.6211 | 13.1123 | 0.0023 |

MSE = 275.77    AFER = 0.658643 %

**Comparison of Proposed Method [Arul], Mr. Jillani and Mr. Lee**

| Year | Actual Values | Jillani | Lee | Arul |
|---|---|---|---|---|
| 2004 | 953 | 995 | 1000 | 1036 |
| 2003 | 1035 | 995 | 1000 | 1036 |
| 2002 | 1145 | 1095 | 1100 | 1149 |
| 2001 | 1288 | 1296 | 1300 | 1280 |
| 2000 | 1253 | 1296 | 1300 | 1249 |
| 1999 | 1173 | 1196 | 1200 | 1182 |
| 1998 | 1224 | 1196 | 1200 | 1216 |
| 1997 | 1150 | 1196 | 1200 | 1149 |
| 1996 | 1122 | 1095 | 1100 | 1092 |
| 1995 | 1228 | 1396 | 1400 | 1216 |
| 1994 | 1415 | 1296 | 1300 | 1415 |
| 1993 | 1346 | 1396 | 1400 | 1338 |
| 1992 | 1380 | 1497 | 1500 | 1384 |
| 1991 | 1471 | 1497 | 1500 | 1476 |
| 1990 | 1574 | 1497 | 1500 | 1577 |
| 1989 | 1488 | 1396 | 1400 | 1493 |
| 1988 | 1432 | 1396 | 1400 | 1430 |
| 1987 | 1390 | 1497 | 1500 | 1384 |
| 1986 | 1456 | 1296 | 1300 | 1461 |
| 1985 | 1308 | 1396 | 1400 | 1305 |
| 1984 | 1369 | 1497 | 1500 | 1369 |
| 1983 | 1479 | 1497 | 1500 | 1476 |
| 1982 | 1464 | 1497 | 1500 | 1461 |
| 1981 | 1564 | 1497 | 1500 | 1555 |
| 1980 | 1616 | 1497 | 1500 | 1622 |
| 1979 | 1572 | 1497 | 1500 | 1577 |
| 1978 | 1644 | 1497 | 1500 | 1644 |
| 1977 | 1597 | 1497 | 1500 | 1599 |
| 1976 | 1536 | 1497 | 1500 | 1533 |
| 1975 | 1460 | 1497 | 1500 | 1461 |
| 1974 | 1574 | 1497 | 1500 | 1577 |

| Measurement | Jilani | Lee | Proposed |
|---|---|---|---|
| MSE | 6908.61 | 6850.38 | 275.77 |
| AFER | 5.061793 | 5.067887 | 0.658643 |

**Figure 1 - Comparison of proposed and other Methods**

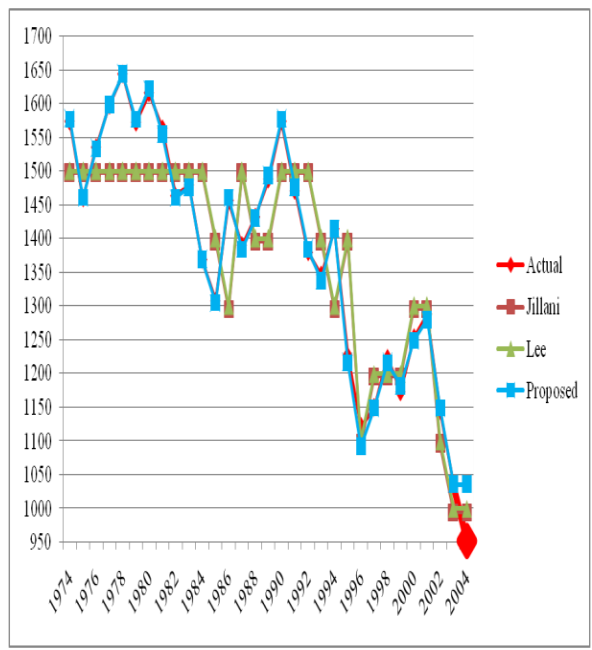

## V. CONCLUSIONS

In this paper, we have presented means based partitioning of the historical accident data of the Belgium. The proposed methods belong to the first order and time-variant methods. From Table VI, we can see that the AFER and MSE of the forecasting results of the proposed method are the smallest than that of the existing method to deal with other forecasting problems based on fuzzy time series. We will also develop new methods for forecasting data based on different intervals from mean parametric approaches to get a higher forecasting accuracy.

## AUTHORS PROFILE

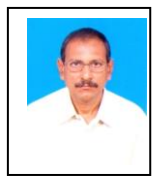

**Arutchelvan Govindarajan** received B.Sc degree in Mathematics from University of Madras, Chennai, M.Sc., degree in Mathematics from Bharathidasan University, Trichy, M.Ed., degree in Mathematics from Annamalai University, Chidambaram, M.C.A., degree in Computer Applications from Madurai Kamaraj University, Madurai, M.Phil., degree in Computer Science from Manonmaniam Sundaranar University, Tirunelveli, PGDCA in Computer Application &

DSADP in Computer Science from Annamalai University, Chidambaram. Presently He is doing Ph.D in Computer Science at SCSVMV University, Kancheepuram. He is the Vice-Principal & Head of the Department of Computer Science and Applications at Adhiparasakthi College of Arts & Science, G.B.Nagar, Kalavai, Vellore Dist., Tamil Nadu . He has 22 years teaching experience in Computer Science and Applications. His research interest is in the area of Fuzzy Logic, Artificial Intelligence, Artificial Neural Networks. He is also a member of Computer Society of India.

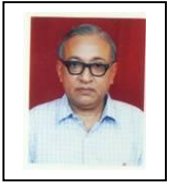

**Dr. S.K.Srivatsa**
Senior professor, St. Joseph's College of Engg,
Jeppiaar Nagar, Chennai-600 064

**Dr.S.K.Srivatsa** was born at Bangalore on 21stJuly 1945. He received his Bachelor of Electronics and Communication Engineering Degree (Honors) from Javadpur University (Securing First Rank and Two Medals). Master Degree in Electrical Communication Engineering (With Distinction) from Indian Institute of Science and Ph.D also from Indian Institute of Science, Bangalore. In July 2005, he retired as professor of Electronics Engineering from Anna University. He has taught twenty-two different courses at the U.G. Level and 40 different courses at P.G. Level during the last 32 years. He has functioned as a Member of the Board of Studies in some Educational Institutions. His name is included in the Computer Society of India database of Resource Professionals. He has received about a dozen awards. He has produced 23 PhDs. He is the author of well over 350 publications.

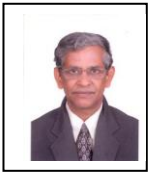

**Dr.Jagannathan Ramaswamy** received B.Sc, M.Sc and Ph.D degrees from the University of Madras, India He obtained his Master of Philosophy degree in Space Physics from Anna University, Chennai. He was the Reader and the Head of the Postgraduate Department of Physics at D.G.Vaishnav College, Chennai.

**Dr.Jagannathan** is currently the Chairman cum Secretary of India Society of Engineers, Madras Chapter, Managing Editor (Publications), Asian Journal of Physics and Deputy Registrar (Education), Vinayaka Missions University, Chennai, India.